\documentclass[runningheads]{llncs}
\usepackage[T1]{fontenc}
\usepackage{graphicx}
\usepackage{amsmath}
\usepackage{algorithm}
\usepackage{algpseudocode}
\begin{document}
\title{A Hybrid Path Planning Framework for Assistive Wheelchair Navigation in Dynamic Pedestrian Environments}
%
%
\author{Damla Leblebicio\u{g}lu\inst{1,2}\orcidID{0000-0002-6288-2980} \and
Rachel Barume\inst{1}\orcidID{0009-0002-6202-1096} \and
Ta\c{s}k{\i}n Pad{\i}r\inst{1,2}\orcidID{0000-0001-5123-5801}}
\authorrunning{D. Leblebicio\u{g}lu et al.}
%
\institute{Electrical and Computer Engineering Department, Northeastern University, Boston, MA, USA \and
Institute for Experiential Robotics, Northeastern University, Boston, MA, USA}
\maketitle              
\begin{abstract}
Autonomous wheelchairs can improve the quality of life for individuals with mobility impairments by providing greater independence. However, reliable navigation especially in human environments is a major challenge due to factors such as dynamic obstacles, unpredictable pedestrian behavior, and the users’ need to complete multiple stops along the way before reaching their final destination. This paper presents a hybrid path planning framework for multi-goal navigation in dynamic environments. The framework integrates the Traveling Salesman Problem (TSP) for global task ordering with the Potential Field (PF) method and heuristic local planners such as Rapidly-exploring Random Tree (RRT) and A-star (A*) to achieve collision-free navigation. Within this hybrid framework, TSP determines the visiting order of intermediate stations, PF transfers global path information, and local planners avoid obstacles while guiding the robot to the stations. The local planners generate feasible paths that allow the robot to follow the global trajectory and reroute to avoid static and dynamic obstacles. For this purpose, two algorithms have been developed: (i) TSP-PF-RRT and (ii) TSP-PF-A*. The performance of the proposed algorithms are evaluated using Monte Carlo simulations in crowded and extra-crowded scenarios that contain heterogeneous dynamic and static obstacles. Experimental results demonstrate that the proposed hybrid framework enables efficient multi-goal navigation while maintaining safe operation in dynamic environments with humans, with TSP–PF–RRT achieving fewer or no collisions and shorter travel distances.


\end{abstract}
\section{Introduction}
Individuals with disabilities often depend on wheelchairs for mobility in their everyday lives. Manual wheelchairs have served a vast majority of these individuals for centuries, yet they often cause strain on the user’s arms and limbs, restricting them to short distances traveled. The introduction of autonomous wheelchairs has granted a large degree of independence for many individuals by enabling the opportunity for safer and longer-distance travel without taking a toll on the users’ bodies or requiring assistance from caregivers. 	

Despite these advances, achieving reliable autonomous navigation in complex environments remains a significant challenge~\cite{khatib1986real,fox1997dynamic}. Path planning for autonomous robots such as autonomous wheelchairs have been studied for decades~\cite{thrun2005probabilistic}; however, existing studies have yet to demonstrate a path planning algorithm that prioritizes dynamic obstacle avoidance, safety constraints, and multi-goal navigation in robust and densely populated environments~\cite{lavalle1998rapidly}. It is critical that autonomous wheelchairs can operate safely and smoothly around pedestrians while simultaneously maintaining appropriate interaction distances. Path planning methods must be able to manage local
obstacle avoidance while balancing global routing efficiency, which ensures that
the system can operate safely within human-centered environments.

Advances in computational power and the decreasing cost of sensing technologies such as LiDAR and stereo cameras have significantly expanded the capabilities of mobile robots operating in dynamic environments~\cite{thrun2005probabilistic,cadena2016past}. These developments enable robots to perceive their surroundings in real-time and support the deployment of computationally demanding algorithms directly on onboard systems. As a result, real-time path planning algorithms that leverage rich sensory information have become increasingly practical for navigating complex and dynamic environments~\cite{paden2016survey}. The work presented in this paper builds upon these technological advances by developing a path planning framework that can operate in real-time using sensor-based obstacle detection. 

Early work on robot path planning has largely relied on potential field (PF) approaches, which have been widely studied for single-goal navigation problems~\cite{khatib1986real}. While PF-based methods can be effective in relatively simple environments, they often struggle in complex scenarios involving multiple goals and dynamic obstacles. In such settings, local minima and the lack of global planning capabilities limit their performance~\cite{guan2025improved,hao2023uav}. Furthermore, achieving reliable obstacle avoidance typically requires careful tuning of PF parameters, which can be difficult and time-consuming. Over the past decade, research has increasingly focused on graph-based and sampling-based planners such as A-star (A*) and Rapidly-exploring Random Trees (RRT), which have demonstrated strong performance in a wide range of robotic navigation tasks~\cite{huang2025self,huang2022research,zhang2018path}. These algorithms are particularly effective at computing feasible paths in cluttered environments and are widely used as core components of modern motion planning systems. In recent years, data-driven approaches have also shown promising results for navigation and decision-making tasks. However, reliable path planning remains essential in safety-critical robotic systems where collision avoidance must be guaranteed~\cite{chehelgami2023safe}.


When the robot must visit multiple task locations, however, the problem becomes significantly more complex. A common approach to addressing multi-goal navigation is to determine the optimal visiting order of goals, which can be formulated as a Traveling Salesman Problem (TSP)~\cite{guo2020global,janovs2021multi,faigl2019fast}. By computing an efficient global visiting sequence, the robot can minimize the overall travel distance while completing multiple tasks. However, when dynamic or randomly moving obstacles are present in the environment, the TSP, which functions as a global path planner (GPP), may require an additional local planning mechanism to enable effective obstacle avoidance during navigation. Despite their strengths, planners such as A* and RRT face challenges in highly dynamic and complex environments. As environment size and complexity increases, the computational burden of these algorithms grows, which can hinder their ability to operate reliably in real-time settings~\cite{huang2025self,wang2020neural,guan2025improved,zhang2018path}. Moreover, these planners are inherently designed for local path generation rather than global multi-goal coordination.

For aforementioned reasons, a practical strategy is to combine global path planning methods such as TSP with local motion planners such as RRT or A*. In such hybrid frameworks, TSP determines the global ordering of tasks, while heuristic planners can handle local navigation and obstacle avoidance. However, an important challenge in these approaches is how to effectively transfer the global information generated by the TSP planner to the heuristic local planners (HLPs) operating in dynamic environments.

In this article, as indicated above, a hybrid strategy that combines two approaches is applied and compared. TSP is used as the GPP, and information about the optimal path is transferred to HLPs using the PF technique. Local path planning, obstacle avoidance, and tracking of the optimal path are performed by HLPs such as A* and RRT. Although other alternatives could have been considered, these methods were selected because they are among the most widely used approaches and demonstrate strong local performance with relatively low computational complexity, especially in small regions. To the best of our knowledge, this is the first study in the literature to convey global path information to local planners using the PF technique, combining a GPP such as TSP with HLPs such as RRT and A* algorithms.

The proposed algorithm has many potential applications. For example, consider a patient using a powered wheelchair enters a hospital cafeteria during lunch hour. The user wishes to visit the salad bar, hot entrée station, beverage counter, and checkout register before settling at an available table. The order in which these stations are visited is non-trivial, backtracking wastes energy and increases exposure to the crowd. Meanwhile, the cafeteria floor is populated with ambulatory patients, nurses, and visitors moving unpredictably between tables and the serving line. The wheelchair must plan an efficient multi-stop route while continuously avoiding moving pedestrians and respecting the narrow clearances between tables.

The remainder of the paper is organized as follows: Section 2 presents the proposed composite path planning methods and their related details; Section 3 describes the simulation environment and the dynamic obstacles; Section 4 presents the simulation results and their discussion; and Section 5 concludes the paper with directions for future work.


\section{Proposed-Path Planning Framework}
\subsection{Problem Formulation}

Path planning methods are generally categorized into global and local planning. Global planning computes an optimal path in a known static environment, whereas local planning focuses on real-time collision avoidance in the presence of dynamic obstacles while guiding the robot toward the goal,~\cite{huang2025self}.

In this work, the global planner is designed from the perspective of a wheelchair user navigating an urban environment. The wheelchair starts from an initial location and aims to reach a final destination while visiting several intermediate stations (waypoints) in a nursing home or hospital cafeteria, such as a salad bar, hot entrée station, or a beverage counter. Each way-point must be visited exactly once before reaching the final goal. The global planner therefore determines the shortest path that connects the start position, all waypoints, and the goal while avoiding large static obstacles in the environment, represented as rectangular regions (i.e., stations in the nursing home or hospital cafeteria). Dynamic obstacles (i.e., pedestrians) are modeled as elliptical shapes and are not considered by the global planner; instead, they are handled by the local planner.

Setting up the environmental model is a fundamental step in wheelchair path planning, as it forms the basis of the algorithmic framework used to safely navigate through a highly dynamic environment without collisions. Consider a wheelchair operating in a 2D environment containing a set of task locations that must be visited before reaching the goal position, $q_{\text{goal}}$. Let $W = \{w_1, w_2, \ldots, w_n\}$ denote the set of $n$ stations, and let $q_0$ represent the initial position of the wheelchair. The global planner determines the optimal visiting order of these stations while minimizing the overall path length, whereas the local planner ensures collision avoidance and guides the robot to follow the planned global path. This formulation naturally becomes a robotic variant of the TSP, where given a set of locations (or cities), the goal is to determine the optimal visiting order that minimizes the total path length while visiting each location exactly once before reaching the goal,~\cite{faigl2019fast}. 

\subsection{Path Planning Environment}
The environment contains heterogeneous obstacles, including rectangular static obstacles and elliptical dynamic obstacles representing pedestrians and other wheelchairs (i.e., in a nursing home), which are modeled with different sizes to reflect their varying physical dimensions.

\subsection{Global Geometric Path Construction}

\begin{figure}
\centering
\includegraphics[width=\textwidth]{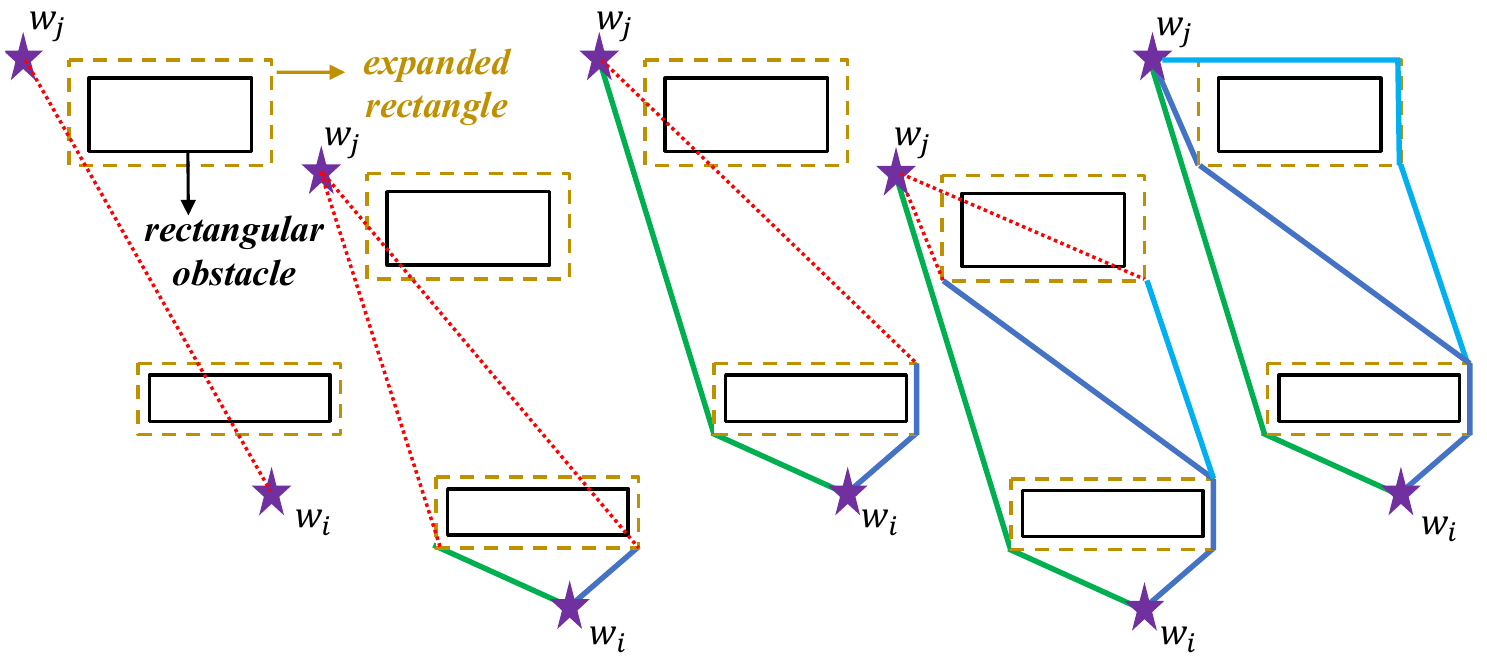}
\caption{Construction of all feasible connections between the start, waypoints, and the goal before solving the TSP.} 
\label{fig:TSP1}
\end{figure}

Determining a path that visits each waypoint exactly once while minimizing the total travel distance corresponds to the classical TSP. Due to the availability of efficient solution methods, TSP is employed to compute the global path,~\cite{faigl2019fast}. To ensure safe navigation, each rectangular obstacle is expanded by a safety margin of 1~m in all directions, forming an expanded rectangle that prevents the robot from entering the obstacle region (illustrated as yellow rectangles in Fig.~\ref{fig:TSP1}). 

A* can also be used as a global planner due to its simplicity and effectiveness in solving shortest-path problems. When the number of waypoints is relatively small, A* can produce the same global path as the TSP-based approach. However, its computational cost increases significantly as the number of waypoints grows, making it impractical for large waypoint sets~\cite{huang2025self}. Therefore, TSP is selected as the global path planner in this study.

The objective of the global planner is to construct all feasible geometric connections between the start position, waypoints, and the goal while avoiding the actual big obstacles (illustrated as black rectangles in Fig.~\ref{fig:TSP1}). Among these connections, the shortest path that visits all waypoints is determined using particle swarm optimization (PSO). The global path is recomputed every 10~s to account for possible environmental changes, for example when rectangular obstacles represent dynamic objects such as large vehicles or trucks.

\begin{figure}
\centering
\includegraphics[width=\textwidth]{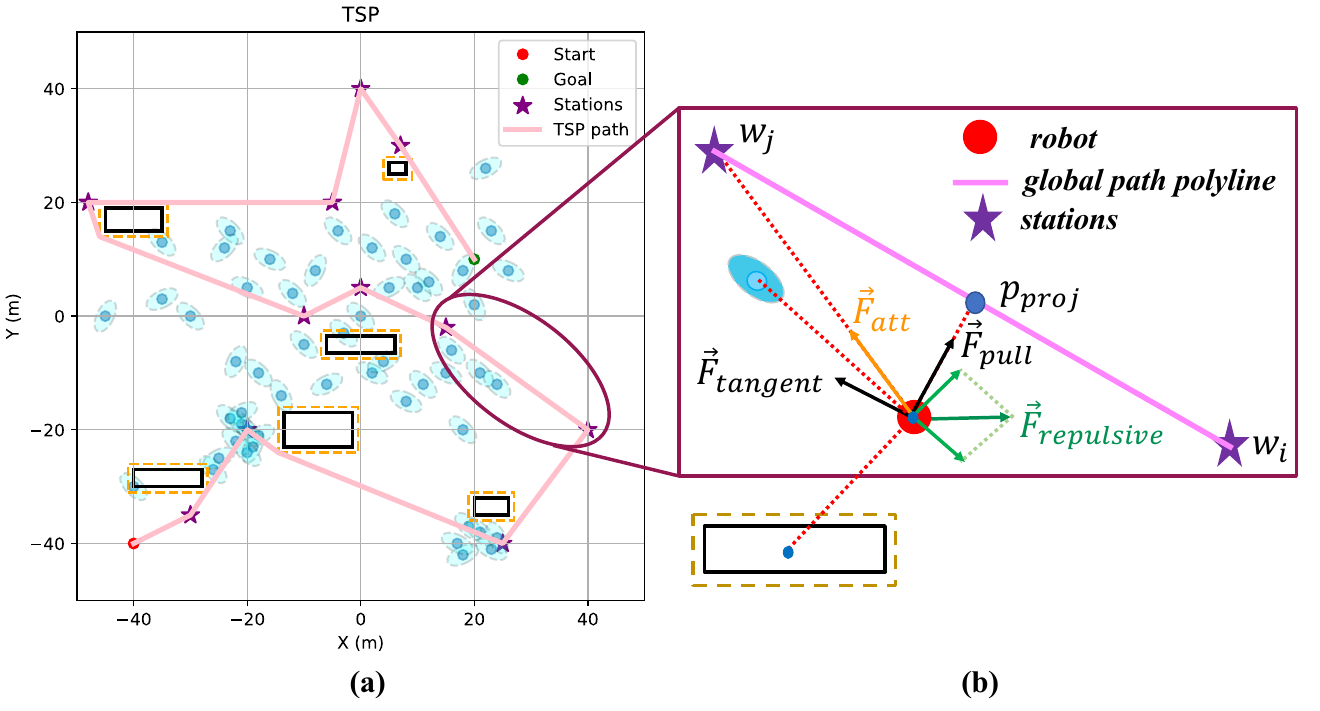}
\caption{(a) Global optimal path for an 11-waypoint problem obtained using PSO, starting from the initial position, visiting each station once, and finally reaching the goal, (b) PF components used for navigation between stations $w_i$ and $w_j$: $\mathbf{F}_{\text{tangent}}$ and $\mathbf{F}_{\text{pull}}$ guide the robot along the global path, $\mathbf{F}_{\text{attractive}}$ pulls the robot toward the next station, and $\mathbf{F}_{\text{repulsive}}$ represents repulsive forces generated by rectangular and elliptical obstacles.}
\label{fig:TSP_path}
\end{figure}

The GPP is constructed as follows:
\begin{enumerate}
    \item Let $V = \{q_0, w_1, w_2, \ldots, w_N, q_{\text{goal}}\}$
    denotes the set of nodes consisting of the start position, waypoints, and the goal. For each pair of nodes from $V$, a direct geometric connection is first considered by drawing a straight line between them (i.e., red line between stations $w_i$ and $w_j$ in Fig.~\ref{fig:TSP1}). If the line does not intersect any rectangular obstacle, the connection is accepted.
    \item If the straight line intersects a rectangular obstacle, the algorithm identifies the closest intersecting rectangle and redirects the path toward all corners of the corresponding expanded rectangle that do not intersect the obstacle, thereby generating all feasible connections.
    \item During this process, visited corners are stored and not revisited again to avoid redundant paths and cycles.
    \item At each step, the algorithm also checks whether a direct collision-free connection to the goal exists from the current node. If such a connection is available, the path of that node is completed (green path in Fig.~\ref{fig:TSP1}); otherwise, search continues along the feasible corners of the expanded rectangles (blue paths in Fig.~\ref{fig:TSP1}).
    \item After constructing all feasible geometric connections between the start, waypoints, and goal, a topological graph is formed. PSO is then applied to solve the resultant TSP. The global path obtained using PSO for an 11-waypoint problem is shown in Fig.~\ref{fig:TSP_path}a.
\end{enumerate}

\subsection{PF Guidance}
PF is used together with HLPs: RRT or A* for local path planning of the robot (wheelchair). The PF provides a directional guidance force, while RRT or A* ensure collision-free motion by locally planning feasible paths. In the PF framework, the robot is attracted toward the goal with lower potential and repelled from obstacles with higher potential through virtual attractive ($\mathbf{F}_{\text{attractive}}$) and repulsive forces ($\mathbf{F}_{\text{repulsive}}$) generated by the potential field,~\cite{guan2025improved}. The formulation for calculating $\mathbf{F}_{\text{attractive}}$ towards stations is given in Eq.~\ref{eq:attractive}, where $k_{att}$ is the coefficient of attraction and $\mathbf{q} = [q_x, q_y]$ is the current position of the robot. 


\begin{equation}
\label{eq:attractive}
\mathbf{F}_{att} = 
k_{att}\frac{\mathbf{q}_{goal}-\mathbf{q}}
{\|\mathbf{q}_{goal}-\mathbf{q}\|}
\end{equation}

Static large obstacles are modeled with rectangles. For rectangular obstacles, a repulsive force is generated when the robot enters a predefined influence distance $d_0$. Each rectangle is approximated by its circumcircle with center $\mathbf{c}_i$ and radius $r_i$. The effective distance between the robot position $\mathbf{q}$ and the rectangle center is $ d_E = \|\mathbf{q} - \mathbf{c}_i\| - r_i$. When $d_E < d_0$, the repulsive force applied by a rectangle is computed as in Eq.~\ref{eq:Frect}, where $k_{rep}$ is the repulsion coefficient. The total repulsive force generated by all rectangular obstacles is obtained by summing the individual forces $\mathbf{F}_{rep}^{rect} = \sum_i \mathbf{F}_{rep}^{(i)}$,~\cite{guan2025improved}.

\begin{equation}
\label{eq:Frect}
\mathbf{F}_{rep}^{(i)} =
k_{rep}
\left(\frac{1}{d_E}-\frac{1}{d_0}\right)
\frac{1}{d_E^2}
\frac{\mathbf{q}-\mathbf{c}_i}{d_E}
\end{equation}

Dynamic obstacles are modeled as ellipses. The distance between the robot position $\mathbf{q}$ and the obstacle center $\mathbf{o}_{i}$ is denoted by $d_E$. When $d_E < d_0$, the obstacle generates a repulsive force acting on the robot (Eq.~\ref{eq:Fellipse}). The magnitude of this force is defined as a piecewise constant value. In the outer part of the influence region, the force magnitude is kept small to allow the robot to smoothly maneuver around the obstacle ($k_{low}$). When the robot approaches the a safety threshold $d_{\text{safe}}$ close to obstacle boundary, the force magnitude switches to a significantly larger constant value ($k_{high}$) in order to strongly repel the robot and prevent collisions. The direction of the repulsive force is always defined along the vector from the obstacle center to the robot position. The repulsive force generated by all elliptical obstacles is obtained by summing the individual forces $\mathbf{F}_{rep}^{ellipse} = \sum_i \mathbf{F}_{rep}^{(i)}$. The total repulsive force acting on the robot is $\mathbf{F}_{\text{repulsive}} = \mathbf{F}_{\text{rep}}^{\text{rect}} + \mathbf{F}_{\text{rep}}^{\text{ellipse}}$.

\begin{equation}
\mathbf{F}_{rep}^{i} =
\begin{cases}
k_{low} \dfrac{\mathbf{q}-\mathbf{o}_{i}}{\|\mathbf{q}-\mathbf{o}_{i}\|}, & d_{safe} < d_E < d_0 \\[6pt]
k_{high} \dfrac{\mathbf{q}-\mathbf{o}_{i}}{\|\mathbf{q}-\mathbf{o}_{i}\|}, & d_E \le d_{safe} \\[6pt]
0, & d_E \ge d_0
\end{cases}
\label{eq:Fellipse}
\end{equation}

There are two additional forces that are associated with the global path. These are the tangent force along the global path, $\mathbf{F}_{\text{tangent}}$, and a pulling force, $\mathbf{F}_{\text{pull}}$, which ensure that the robot follows the global path even when it is disturbed by the repulsive forces generated by obstacles ($\mathbf{F}_{\text{repulsive}}$). In Eq.~\ref{eq:Fglobal}, $\mathbf{p}_{\text{proj}}$ denotes the closest point to the robot position $\mathbf{q}$ on the global path polyline, and $\hat{\mathbf{t}}$ denotes the unit tangent vector of the corresponding path segment, $k_{pull}$ and $k_{tangent}$ are attractive force coefficients towards the global path. The total force acting on the robot is $\mathbf{F}_{\text{total}} =
\mathbf{F}_{\text{attractive}} +
\mathbf{F}_{\text{repulsive}} +
\mathbf{F}_{\text{tangent}} +
\mathbf{F}_{\text{pull}}$, shown in Fig.~\ref{fig:TSP_path}b. 

\begin{equation}
\label{eq:Fglobal}
\mathbf{F}_{\text{pull}} = k_{\text{pull}} (\mathbf{p}_{\text{proj}} - \mathbf{q}), \qquad
\mathbf{F}_{\text{tangent}} = k_{\text{tan}} \hat{\mathbf{t}}
\end{equation}

\subsection{Local Collision-Free Planning with RRT}
\label{sec:RRT}
In the first approach, RRT is used as the HLP together with the PF method in the proposed hybrid planner. RRT is a probabilistic algorithm that incrementally builds a tree from the start toward the goal by randomly sampling points in the workspace. For each sampled point, the nearest node in the existing tree is identified and the tree is extended toward the sampled point if the path segment is collision-free. This process is repeated until the goal is reached. Unlike grid-based search methods such as A*, RRT does not require modeling the entire configuration space explicitly. Instead, it incrementally explores the state space by expanding a tree from the start toward the goal through random sampling, allowing it to efficiently solve path planning problems in complex environments~\cite{zhang2018path}, particularly when compared to grid-based planners such as A*.


If the start or goal node of the RRT lies within a rectangular or elliptical obstacle, it is displaced in the direction of the $\mathbf{F}_{\text{repulsive}}$ using the RRT step size until a collision-free position is obtained. The direction of $\mathbf{F}_{\text{total}}$ provides guidance for determining the local target of the RRT. The local goal for the RRT expansion is defined within a local horizon $L$ in the direction of $\mathbf{F}_{\text{total}}$ and is added to the current robot position $\mathbf{q}$, as given in Eq.~\ref{eq:RRT}.

\begin{equation}
\label{eq:RRT}
\mathbf{q}_{\textit{target\_local}} =
\mathbf{q} + L
\frac{\mathbf{F}_{total}}
{\|\mathbf{F}_{total}\|}
\end{equation}

\subsection{Local Collision-Free Planning with A*}
In the second approach, A* is used as the HLP together with the PF method in the proposed hybrid planner. A* is a heuristic search algorithm designed to find the global optimal path in environments with static obstacles. The method is grid-based, where the environment is divided into small square cells (grids), and the optimal path is determined by evaluating the total cost function $f(n)$ for neighboring nodes while progressing toward the goal. 

The A* algorithm maintains two values to calculate the total cost $f(n)$: $g(n)$ and $h(n)$, where $n$ represents the current node in the search. The term $g(n)$ denotes the cost from the start node to the current node, while $h(n)$ represents the heuristic estimate of the cost from the current node to the goal. The total cost function is defined in Eq.~\ref{eq:Astar}. During the search process, the node with the smallest total cost $f(n)$ is selected from the open set and expanded until the goal node is reached,~\cite{zhang2023global}. Neighbor nodes that lie inside static rectangular obstacles are ignored. Similar to the local planner RRT implementation described in Section~\ref{sec:RRT}, a local target ($\mathbf{q}_{\textit{target\_local}}$) is generated within a local horizon ($L$) in the direction of $\mathbf{F}_{\text{total}}$. A* plans a path from the current robot position to the local target while restricting the search to the local region within limited bounds.

\begin{equation}
\label{eq:Astar}
f(n) = g(n) + h(n)
\end{equation}


\subsection{Hybrid Path Construction for Multi-Goal Navigation}
\label{sec:hybrid}

In dynamic environments, global and local planning are typically handled separately. In the proposed hybrid framework, the global planning problem is formulated as a TSP and solved using PSO. At the local level, robot motion is guided by the PF method together with RRT or A* to avoid collisions with both static and dynamic obstacles. Algorithm~\ref{alg:hybrid} presents the pseudocode for the two proposed approaches: (i) TSP-PF-RRT and (ii) TSP-PF-A*. 

\begin{algorithm}[t]
\caption{Hybrid TSP–PF–RRT or TSP–PF–A* Planner}
\label{alg:hybrid}
\begin{algorithmic}[1]

\State Solve TSP using PSO to obtain the global waypoint order
\State Construct the global geometric path $\mathcal{G}$ and polyline
\State $\mathbf{q}_{goal} \gets$ next waypoint on $\mathcal{G}$

\While{$\mathbf{q} \neq \mathbf{q}_{goal\_final}$}

\State Apply stochastic maneuver to obstacle velocities
\State Update obstacle positions
\State Compute $\mathbf{F}_{total}$ using $\mathbf{F}_{attractive}$, $\mathbf{F}_{repulsive}$, $\mathbf{F}_{tangent}$ and $\mathbf{F}_{pull}$
\State $\mathbf{d} \gets \mathbf{F}_{total}/\|\mathbf{F}_{total}\|$
\State $\mathbf{q}_{\textit{target\_local}} \gets \mathbf{q} + L\mathbf{d}$
\State Plan path from $\mathbf{q}$ to $\mathbf{q}_{\textit{target\_local}}$ using RRT or A*
\State $\mathbf{q}_{next} \gets$ next node of the RRT or A* path
\State $\mathbf{q} \gets \mathbf{q} + v_{robot} dt \frac{\mathbf{q}_{next}-\mathbf{q}}{\|\mathbf{q}_{next}-\mathbf{q}\|}$
\State Check collisions with obstacles
\If{current waypoint reached}
    \State $\mathbf{q}_{goal} \gets$ next waypoint
\EndIf

\EndWhile

\end{algorithmic}
\end{algorithm}

\section{Simulation Environment}
The environment is defined over the range $[-50,50]$ m in both the $x$ and $y$ directions. The initial position of the wheelchair is $\mathbf{q}_0 = [-40,-40]$, and the goal position is $\mathbf{q}_{\text{goal}} = [20,10]$. For the dynamic obstacles, ellipse parameters are scaled according to the velocity magnitude. The major axis $a$ is aligned with the direction of velocity, while the minor axis $b$ is perpendicular to the direction of velocity. The lengths of the axes are defined in Eq.~\ref{eq:a_b}, where $\|\mathbf{v}\|$ is the velocity magnitude, and $\alpha = 0.2$ and $\beta = 0.1$ are scaling coefficients. The maximum values of the axes are limited to $a_{\max} = 3\,\mathrm{m}$ and $b_{\max} = 1.5\,\mathrm{m}$. The simulation time step is $\Delta t = 0.01\,\mathrm{s}$. We also introduce stochastic maneuvering behavior for dynamic obstacles. Specifically, they may exhibit small speed fluctuations (i.e., slight increases or decreases in speed), large random maneuvers, or small random drifts in the direction of their velocity.

\begin{equation}
\label{eq:a_b}
a = a_{\text{base}} + \alpha \|\mathbf{v}\|, \qquad
b = b_{\text{base}} + \beta \|\mathbf{v}\|
\end{equation}

To simulate sensing uncertainty, Gaussian noise with zero mean and a standard deviation of $\sigma = 0.1~\mathrm{m}$ is added to the elliptical obstacle positions. The perceived obstacle positions are modeled in Eq.~\ref{eq:obstacle_noise}, where $\mathbf{p}_{\mathrm{obs}}^{\mathrm{true}}$  and $\mathbf{p}_{\mathrm{obs}}^{\mathrm{noisy}}$ denote the true and noisy obstacle positions, respectively.

\begin{equation}
\label{eq:obstacle_noise}
\mathbf{p}_{\mathrm{obs}}^{\mathrm{noisy}} = \mathbf{p}_{\mathrm{obs}}^{\mathrm{true}} + \boldsymbol{\epsilon}, 
\qquad 
\boldsymbol{\epsilon} \sim \mathcal{N}\!\left(\mathbf{0},\, \sigma^2\right)
\end{equation}

\begin{figure}
\centering
\includegraphics[width=0.75\textwidth]{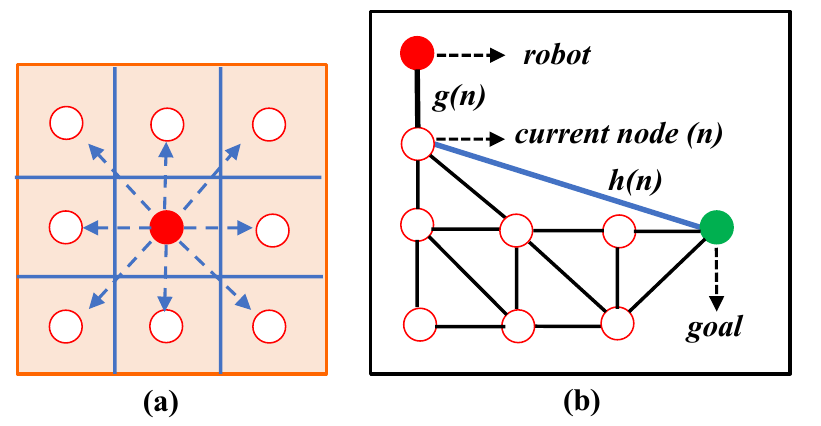}
\caption{(a) Illustration of the 8-neighborhood search used in A*, (b) Computation of the total cost function $f(n)=g(n)+h(n)$.} 
\label{fig:Astar}
\end{figure}

In the simulation, the search space for the A* algorithm is restricted to a local region centered at the current robot position. Specifically, the search bounds are defined as $x \in [q_x-5, q_x+5]$ and $y \in [q_y-5, q_y+5]$, and the region is represented on a grid with a resolution of 0.5 m. Motion is allowed in eight directions (8-neighborhood search): up, down, left, right, upper-left, upper-right, lower-left, and lower-right. Consequently, the cost $g(n)$ of moving to a diagonal successor node is $\sqrt{0.5}$ m, while the cost of moving in the horizontal or vertical directions is 1 m. The heuristic function $h(n)$ is defined as the Euclidean distance between the current node and the goal, as shown in Fig.~\ref{fig:Astar}b.

\section{Simulation Results and Discussion}
\begin{figure}
\centering
\includegraphics[width=\textwidth]{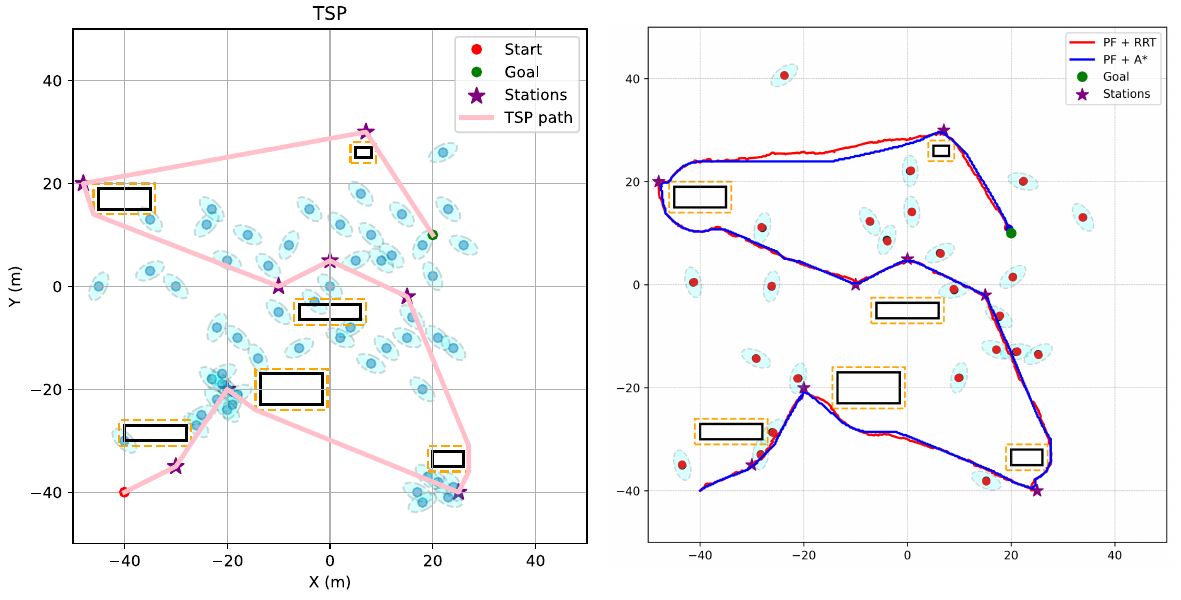}
\caption{Left: global path. Right: robot trajectories generated by (i) PF-RRT (red) and (ii) PF-A* (blue) in a crowded environment.}
\label{fig:crowded}
\end{figure}

\begin{figure}
\centering
\includegraphics[width=\textwidth]{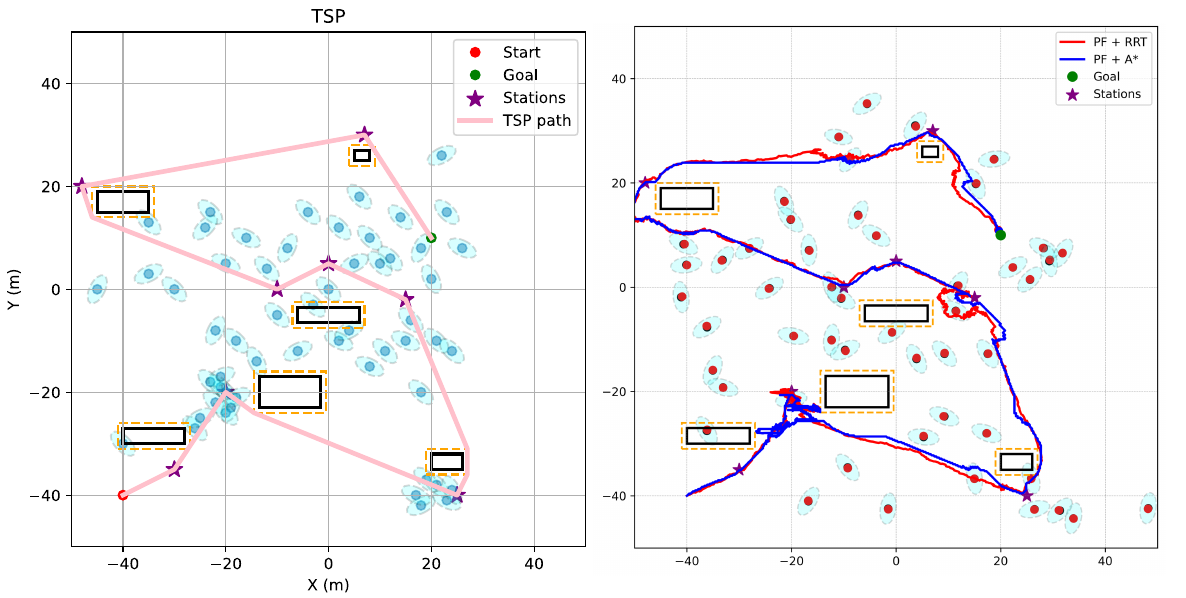}
\caption{Left: global path. Right: robot trajectories generated by (i) PF-RRT (red) and (ii) PF-A* (blue) in an extra-crowded environment.}
\label{fig:extracrowded}
\end{figure}

The PF technique alone was not preferred, as it cannot effectively avoid collisions in crowded environments. The same limitation applies to A* when used solely for local planning. For these reasons, neither method was considered a suitable basis for comparison. In our evaluation, the proposed methods described in Section~\ref{sec:hybrid} are compared through Monte Carlo simulations, where TSP is used for global planning and (i) PF–RRT and (ii) PF–A* are used for local obstacle avoidance. Monte Carlo simulations are employed because obstacle motions are stochastic. Ten simulations are performed under two different scenarios: crowded and extra-crowded. In both scenarios, the two methods are supposed to follow the same global path obtained by PSO for an eight-waypoint trajectory. For each scenario, the average total distance traveled and the number of collisions over the ten trials are reported in Table~\ref{tab:comparison}. In addition, the global path together with the robot trajectory from one representative trial for both methods are shown in Figs.~\ref{fig:crowded} and~\ref{fig:extracrowded}.

As shown in Table~\ref{tab:comparison}, the average total travel distance obtained by both methods is very similar in both scenarios, with a slight advantage for RRT. No collisions occur when using the TSP–PF–RRT method. In contrast, collisions occur under the TSP–PF–A* method, and the number of collisions increases as the environment becomes more crowded. This behavior is expected due to the nature of the two methods. A* is primarily designed for static environments and is relatively simple, whereas RRT tends to perform better as the environment becomes more complex; although, RRT is computationally more demanding. For both methods, the total travel distance naturally increases as the number of obstacles increases, since the robot must perform additional avoidance maneuvers around obstacles. It is important to note that both methods perform well overall. Although TSP–PF–A* exhibits a higher number of collisions, this is mainly due to the large number of moving obstacles in the second scenario, which occasionally trap the robot in confined spaces.

\begin{table}[t]
\centering
\caption{Comparison of the proposed methods over 10 Monte Carlo trials in crowded and extra-crowded environments.}
\label{tab:comparison}

\begin{tabular}{lcccc}
\hline
Method & \multicolumn{2}{c}{Crowded} & \multicolumn{2}{c}{Extra-crowded} \\
\cline{2-5}
 & Distance (m) & Collisions & Distance (m) & Collisions \\
\hline
TSP--PF--RRT & 319 & 0 & 525 & 0 \\
TSP--PF--A*  & 340 & 2.8 & 533.7 & 7.4 \\
\hline
\end{tabular}

\end{table}

\section{Conclusion and Future Work}

In this study, fully automated path planning algorithms capable of performing multi-goal navigation in environments with dynamic obstacles are developed and compared. The primary objective is to enable wheelchair users to autonomously reach their destinations without fatigue while safely navigating among people and avoiding collisions. The proposed algorithms can be applied in both indoor and outdoor environments.
To deploy the proposed framework in real-world applications, several extensions can be considered both algorithmically and in terms of hardware. From an algorithmic perspective, the system could allow users to specify desired destinations on a map, enabling the global path to be planned according to user preferences and waypoint order. Additionally, the environment map could be integrated into the onboard computer of the wheelchair for both indoor and outdoor navigation.
From a hardware perspective, the system could be implemented on a powered wheelchair equipped with sensors such as GNSS, LiDAR, and stereo cameras. Using these sensors, the positions of obstacles can be estimated in real-time, enabling the proposed algorithms to operate in real-world environments.
Furthermore, obstacles could be classified according to their type and intent. Depending on their behavior, different interaction strategies could be applied. For example, pedestrians and other wheelchairs could be treated as different obstacle classes. In some cases, it may be preferable to wait rather than perform an avoidance maneuver. For instance, when another wheelchair temporarily blocks a desired location. Such behavior-aware obstacle handling strategies could further improve navigation safety and efficiency.


\begin{credits}
\subsubsection{\ackname} 
This research was funded, in part, by the Advanced Research Projects Agency for Health (ARPA-H) Agreement No. 140D042590012. The views and conclusions contained in this document are those of the authors and should not be interpreted as representing the official policies, either expressed or implied, of the U.S. Government.

\subsubsection{\discintname}
Ta\c{s}k{\i}n Pad{\i}r holds concurrent appointments as a Professor of Electrical and Computer Engineering at Northeastern University and as an Amazon Scholar. This paper describes work performed at Northeastern University and is not associated with Amazon.
\end{credits}

%
%
%
\bibliographystyle{splncs04}
\bibliography{mybibliography}

@String{Springer = "Springer-Verlag" }

@article{huang2025self,
  title={A self-adaptive neighborhood search A-star algorithm for mobile robots global path planning},
  author={Huang, Jiabo and Chen, Chunmei and Shen, Junjie and Liu, Guihua and Xu, Feng},
  journal={Computers and Electrical Engineering},
  volume={123},
  pages={110018},
  year={2025},
  publisher={Elsevier}
}

@article{zhang2023global,
  title={Global path planning of unmanned surface vehicle based on improved A-Star algorithm},
  author={Zhang, Huixia and Tao, Yadong and Zhu, Wenliang},
  journal={Sensors},
  volume={23},
  number={14},
  pages={6647},
  year={2023},
  publisher={MDPI}
}

@article{faigl2019fast,
  title={Fast heuristics for the 3-D multi-goal path planning based on the generalized traveling salesman problem with neighborhoods},
  author={Faigl, Jan and V{\'a}{\v{n}}a, Petr and Deckerov{\'a}, Jind{\v{r}}i{\v{s}}ka},
  journal={IEEE Robotics and Automation Letters},
  volume={4},
  number={3},
  pages={2439--2446},
  year={2019},
  publisher={IEEE}
}

@article{guan2025improved,
  title={An improved artificial potential field with RRT star algorithm for autonomous vehicle path planning},
  author={Guan, Tian and Han, Yi and Kong, Michuang and Wang, Siyu and Feng, Deyuan and Yang, Wei},
  journal={Scientific Reports},
  volume={15},
  number={1},
  pages={16982},
  year={2025},
  publisher={Nature Publishing Group UK London}
}

@article{zhang2018path,
  title={Path planning of industrial robot based on improved RRT algorithm in complex environments},
  author={Zhang, Haojian and Wang, Yunkuan and Zheng, Jun and Yu, Junzhi},
  journal={IEEE Access},
  volume={6},
  pages={53296--53306},
  year={2018},
  publisher={IEEE}
}

@article{guo2020global,
  title={Global path planning and multi-objective path control for unmanned surface vehicle based on modified particle swarm optimization (PSO) algorithm},
  author={Guo, Xinghai and Ji, Mingjun and Zhao, Ziwei and Wen, Dusu and Zhang, Weidan},
  journal={Ocean Engineering},
  volume={216},
  pages={107693},
  year={2020},
  publisher={Elsevier}
}

@article{janovs2021multi,
  title={Multi-goal path planning using multiple random trees},
  author={Jano{\v{s}}, Jaroslav and Von{\'a}sek, Vojt{\v{e}}ch and P{\v{e}}ni{\v{c}}ka, Robert},
  journal={IEEE Robotics and Automation Letters},
  volume={6},
  number={2},
  pages={4201--4208},
  year={2021},
  publisher={IEEE}
}

@article{hao2023uav,
  title={UAV path planning based on improved artificial potential field method},
  author={Hao, Guoqiang and Lv, Qiang and Huang, Zhen and Zhao, Huanlong and Chen, Wei},
  journal={Aerospace},
  volume={10},
  number={6},
  pages={562},
  year={2023},
  publisher={MDPI}
}

@article{wang2020neural,
  title={Neural RRT*: Learning-based optimal path planning},
  author={Wang, Jiankun and Chi, Wenzheng and Li, Chenming and Wang, Chaoqun and Meng, Max Q-H},
  journal={IEEE Transactions on Automation Science and Engineering},
  volume={17},
  number={4},
  pages={1748--1758},
  year={2020},
  publisher={IEEE}
}

@article{khatib1986real,
  title={Real-time obstacle avoidance for manipulators and mobile robots},
  author={Khatib, Oussama},
  journal={The international journal of robotics research},
  volume={5},
  number={1},
  pages={90--98},
  year={1986},
  publisher={Sage Publications Sage CA: Thousand Oaks, CA}
}

@article{huang2022research,
  title={Research on path planning algorithm of autonomous vehicles based on improved RRT algorithm},
  author={Huang, Guanghao and Ma, Qinglu},
  journal={International journal of intelligent transportation systems research},
  volume={20},
  number={1},
  pages={170--180},
  year={2022},
  publisher={Springer}
}

@article{chehelgami2023safe,
  title={Safe deep learning-based global path planning using a fast collision-free path generator},
  author={Chehelgami, Shirin and Ashtari, Erfan and Basiri, Mohammad Amin and Masouleh, Mehdi Tale and Kalhor, Ahmad},
  journal={Robotics and Autonomous Systems},
  volume={163},
  pages={104384},
  year={2023},
  publisher={Elsevier}
}

@article{fox1997dynamic,
  title={The dynamic window approach to collision avoidance},
  author={Fox, Dieter and Burgard, Wolfram and Thrun, Sebastian},
  journal={IEEE Robotics and Automation Magazine},
  volume={4},
  number={1},
  pages={23--33},
  year={1997}
}

@article{lavalle1998rapidly,
  title={Rapidly-exploring random trees: A new tool for path planning},
  author={LaValle, Steven M},
  journal={Technical Report},
  institution={Iowa State University},
  year={1998}
}

@book{thrun2005probabilistic,
  title={Probabilistic Robotics},
  author={Thrun, Sebastian and Burgard, Wolfram and Fox, Dieter},
  publisher={MIT Press},
  year={2005}
}

@article{cadena2016past,
  title={Past, present, and future of simultaneous localization and mapping: Toward the robust-perception age},
  author={Cadena, Cesar and Carlone, Luca and Carrillo, Henry and Latif, Yasir and Scaramuzza, Davide and Neira, Jose and Reid, Ian and Leonard, John},
  journal={IEEE Transactions on Robotics},
  volume={32},
  number={6},
  pages={1309--1332},
  year={2016}
}

@article{paden2016survey,
  title={A survey of motion planning and control techniques for self-driving urban vehicles},
  author={Paden, Brian and {\v{C}}{\'a}p, Michal and Yong, Sze Zheng and Yershov, Dmitry and Frazzoli, Emilio},
  journal={IEEE Transactions on Intelligent Vehicles},
  volume={1},
  number={1},
  pages={33--55},
  year={2016}
}
%






\end{document}